\documentclass[letterpaper, 10 pt, conference]{ieeeconf}

\IEEEoverridecommandlockouts                              
\usepackage{cite}
\usepackage{amsmath,amssymb,amsfonts}
\usepackage{graphicx}
\usepackage{textcomp}
\usepackage{xcolor}
\def\BibTeX{{\rm B\kern-.05em{\sc i\kern-.025em b}\kern-.08em
    T\kern-.1667em\lower.7ex\hbox{E}\kern-.125emX}}
\usepackage{booktabs}
\usepackage{algorithm}
\usepackage{algpseudocode}
\usepackage{comment}
\usepackage{stfloats}
\usepackage{caption}
\usepackage{subcaption}
\usepackage{blindtext}
\usepackage{hyperref}

\title{\LARGE \bf
Projected Task-Specific Layers for Multi-Task Reinforcement Learning
}

\author{Josselin Somerville Roberts$^{1}$ and Julia Di$^{2}$
\thanks{$^{1}$Josselin Somerville Roberts is with the Department of Computer Science,
        Stanford University, Stanford CA, 94305, USA
        {\tt\small josselin@stanford.edu}}%
\thanks{$^{2}$Julia Di is with the Department of Mechanical Engineering, Stanford University,
        Stanford, CA 94305, USA
        {\tt\small juliadi@stanford.edu}}%
}

\begin{document}

\maketitle

\thispagestyle{empty}
\pagestyle{empty}


\begin{abstract}
Multi-task reinforcement learning could enable robots to scale across a wide variety of manipulation tasks in homes and workplaces. However, generalizing from one task to another and mitigating negative task interference still remains a challenge. Addressing this challenge by successfully sharing information across tasks will depend on how well the structure underlying the tasks is captured. In this work, we introduce our new architecture, Projected Task-Specific Layers (PTSL), that leverages a common policy with dense task-specific corrections through task-specific layers to better express shared and variable task information. We then show that our model outperforms the state of the art on the MT10 and MT50 benchmarks of Meta-World consisting of 10 and 50 goal-conditioned tasks for a Sawyer arm.

\end{abstract}


\section{Introduction}
\label{sec:intro}

\begin{figure*}[b]
  \centering
   \includegraphics[width=1\linewidth]{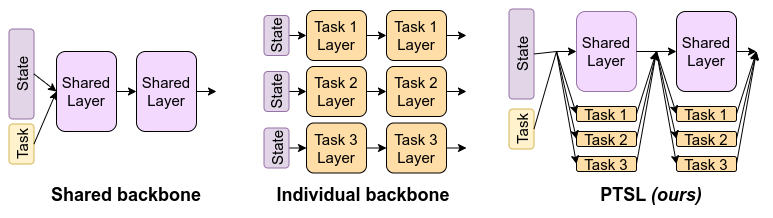}
   \caption{Simplified diagram of different architectures for multi-task reinforcement learning: Shared backbone for all tasks \textit{(left)}, Individual backbone for each task \textit{(center)} and Projected Task-Specific Layers \textbf{(ours)} \textit{(right)}}. 
   \label{fig:methods}
\end{figure*}

Complex manipulation is common in a number of desirable real-world robotic use cases---such as wiping various kitchen surfaces in a busy restaurant, routing cables in a datacenter, or screwing parts in an manufacturing assembly. While these individual tasks are different, they are often composed of similar manipulation primitives. Humans intuitively recognize and scaffold upon these primitives to implement different tasks, but it is challenging for robots, which are often only trained on individual, specific tasks, to do the same. Enabling robots to reason through multiple related tasks with multi-task reinforcement learning~\cite{multitask} can unlock more general-purpose robotics---allowing efficient learning across similar tasks and using their shared structure to learn a better performing policy~\cite{zhang2021survey,yu2020gradient, care}. 

Multi-task learning often, however, requires a careful choice of tasks and balanced sampling, and even then may not always improve learning. For complex manipulation, learning-based approaches may generalize over unseen tasks~\cite{mahler2019learning} but can still be difficult to scale successfully~\cite{kalashnikov2021mt,schaul2019ray}. Recent work has argued that learning methods should use shared task structures~\cite{sodhani2021multi,d2019sharing} but most approaches still learn a single shared policy used across all tasks which may not adequately represent variations between tasks~\cite{kalashnikov2021mt,levine2016end}.

Instead, we propose a new backbone architecture, the Projected Task-Specific Layers (PTSL), which combines a large, shared fully-connected policy with low-rank task-specific layers as shown in Fig.~\ref{fig:methods}. After each layer, the hidden state from the shared policy and the low-rank task-specific policy are combined, making PTSL expressive of different tasks. We evaluate PTSL as a standalone backbone or on top of the Context-based Representation (CARE~\cite{care}) encoder that leverages text descriptions of the task as metadata to project a mixture of encodings.

The main contributions of this work are: 
\begin{itemize}
    \item We propose the \textbf{PTSL} architecture for deep multi-task reinforcement learning that adds low-rank task-specific layers on top of each layer of a shared backbone.
    \item Our results with PTSL outperforms CARE~\cite{care}, the current state-of-the-art, on both the MT10 and MT50 Goal-conditioned benchmarks from Meta-World~\cite{metaworld}. 
    \item Further, our results suggest that multi-task learning with a shared projection is more sample efficient and can improve learning on individual tasks. 
    \item Finally, our results provide insights into the benefits of intermediate architectures sharing an embedding space between task-specific layers and a backbone.
\end{itemize}

\section{Related work}

\subsection{Reinforcement Learning for Robotics}

Reinforcement learning (RL) has recently shown success in various domains such as Atari~\cite{atari}, Go~\cite{alphago}, and Starcraft~\cite{starcraft}. In robotics, RL has been well studied in self-driving cars~\cite{driving}, manipulation~\cite{manipulation,visuomotor,continuous}, and locomotion~\cite{locomotion}. Multi-task reinforcement learning is a subfield of RL for teaching a single agent to solve multiple tasks. Whereas single-task RL optimises for one reward function, multi-task learning optimises for multiple objectives~\cite{zhang2021survey}. 

One of the main challenges of multi-task learning is \textit{task interference} or \textit{negative transfer}: learning a new task can deteriorate the performance of a previously learned task. This has been documented in previous literature~\cite{yu2020gradient, negative_transfer} and several approaches have been proposed for mitigation. One approach is to train single policies for each task and use knowledge transfer~\cite{xu2020knowledge} or policy distillation~\cite{distillation}, but this requires separate networks or a large number of network parameters. Other previous approaches~\cite{pcgrad,vaccine} have addressed this issue by modifying the training algorithm or having a model architecture that can support different sub-policies~\cite{softmod,care}, but these methods are often slow and scale poorly to the number of tasks. Thus, in this work we choose to address this issue by focusing not on the optimization method but the underlying architecture.

\subsection{Soft Actor-Critic}

The Soft Actor-Critic (SAC)~\cite{sac} algorithm optimizes the maximum-entropy RL objective using off-policy data to learn. SAC has been demonstrated to perform better~\cite{metaworld} than other algorithms such as Proximal Policy Optimization (PPO)~\cite{ppo} or Trust Region Policy Optimization (TRPO)~\cite{trpo}. Accordingly, we choose to use the multi-task adaptation of SAC with disentangled alphas (which refers to having separate alpha coefficients for every task learned by the policy~\cite{garage}) to focus on the architecture of the agent.

\subsection{Multi-task Architectures}
Two common and opposing architectures for multi-task learning are the multi-headed actor and the shared actor architecture~\cite{metaworld}. The multi-headed actor consists of a single network with one head per task, but does not scale well due to the sheer number of parameters. Meanwhile, the shared actor consists of having a single network for all tasks, which does not allow for task-specific corrections and often leads to poor performance as the number of tasks increases~\cite{cheng2023multi}.

Within these paradigms, a number of approaches exist. Mixture of Experts (MOE) ~\cite{moe,jacobs1991adaptive} methods have independent experts with a learned gating network to output weights.  Similar work has been done with Hard routing~\cite{hard_routing} which consists of having a routing network that selects the expert for each task. However, good expert assignments can be nontrivial. A third improvement on the multi-headed architecture is called Soft Modularization~\cite{softmod}, where each "step" of the network is composed of multiple linear layers and a routing network decides how to route the linear layers of one step to the ones of the next step. 

Another approach is to have a shared network with a preprocessor that produces a vector to represent the state, often denoted as an encoder. CARE (Contextual Attention-based Representation learning) proposes a mixture of encoder architecture~\cite{care}. The idea is to have several independent encoders that encode the task in a feature vector. Then, another component produces attention scores based only on the task of combining these embeddings. This embedding is then fed to a fully linear network. This process is done for the actor, critic, and value functions. However, while a task-dependent encoder is useful, a shared policy like in CARE~\cite{care} may not be the best approach because some tasks may require small variations in the shared policy.

\section{Method}
\label{sec:method}

In this section, we introduce our \textbf{Projected Task-Specific Layers (PTSL)} architecture, which can be used on top of methods such as CARE\cite{care}. PTSL reconciles encoding approaches and routing approaches for multi-tasking. 

\subsection{Projected Attention Layers}

Encoding the state in a task-specific way, like in CARE, may not be enough for all tasks. For example, in manipulation, knowing how to push and pull may not benefit the actions of picking and placing objects. Yet it is still beneficial to have a shared policy because some skills can be shared.

PTSL permits small task-specific variations in the policy and was inspired by Projected Attention Layers (PAL)~\cite{pal}, a method that permits a transformer to have small task-specific variations for Natural Language Processing (NLP). For each task, the input goes through a series of transformer layers. Each layer is made of two parts: one large shared  attention layer and one small task-specific attention layer. The latter is computed by projecting down the input to a smaller dimension, applying an attention layer trained specifically for the task, and then projecting it up to the original dimension and adding it to the shared attention layer output. This process is repeated for each layer of the transformer. 

The intuition is that a general backbone can contain most of the knowledge (akin to grammar rules for a NLP Transformer), but small task-specific variations may be added as some tasks may require different treatment (e.g. classification versus summarization). PALs have been shown to be very efficient in NLP tasks and we believe that this approach can be applied to robotic manipulation as well.

\begin{figure*}[!t]
    \centering
    \includegraphics[width=\textwidth]{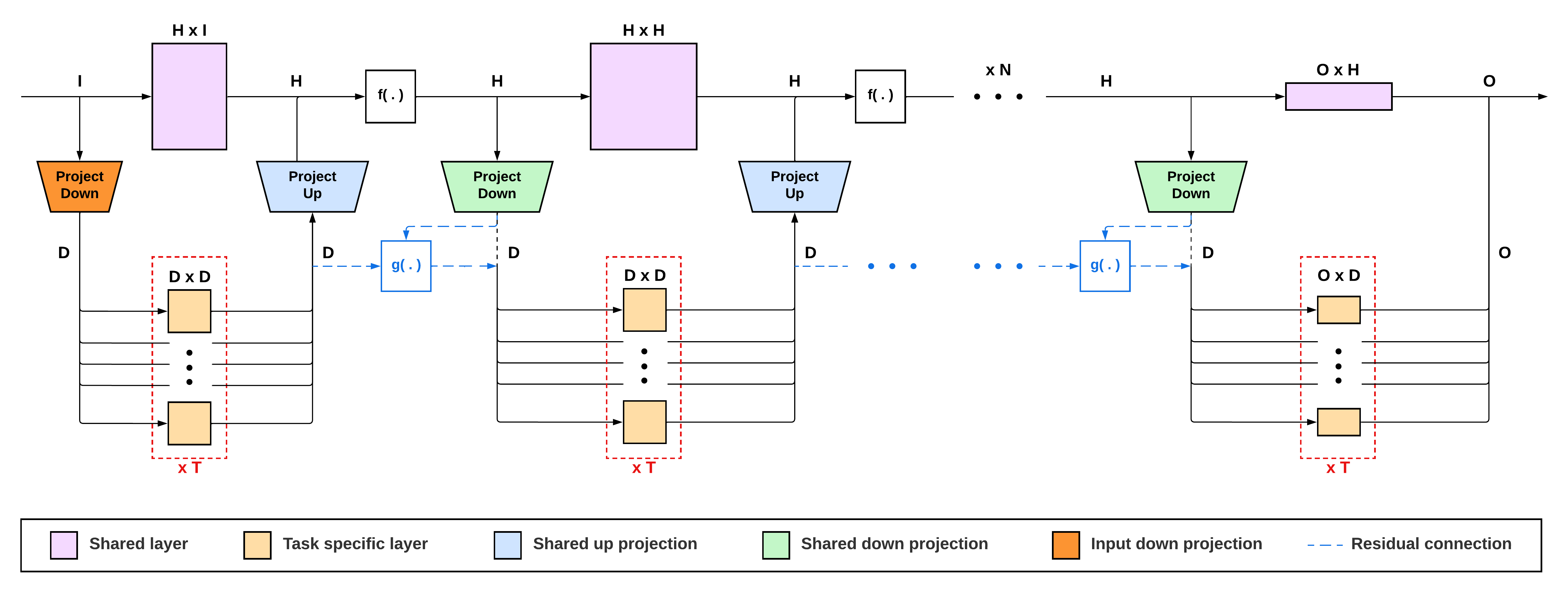}
    \caption{PTSL Architecture \textbf{(ours)}, explained in Section \ref{sec:ptsl} for details. The dotted red lines represent residual connections (not always present). See Section \ref{sec:residuals} for details. Projection modules that are reused are represented with the same color. See Section \ref{sec:projection} for details.}
    \label{fig:ptsl}
\end{figure*}

\subsection{Projected Task-Specific Layers}
\label{sec:ptsl}
We propose \textbf{Projected Task-Specific Layers (PTSL)}, which adapts the PAL architecture to a linear layer setup instead of a transformer, with a few modifications. This architecture can replace the backbone of other methods to obtain complex agents such as \textbf{CARE + PTSL}.

Similarly to PAL, PTSL is made of a shared backbone and low-rank task-specific layers (one low-rank layer for each task, on top of each shared layer) as detailed in Fig.~\ref{fig:ptsl}. The backbone is a linear layer that is shared between all tasks. The task-specific layers are linear layers that are specific to each task. Where our approach differs from PAL is regarding the projections to the task-specific dimension.

\subsection{Problem formulation and Preliminaries}

In a multi-task setting, we have a set of tasks $\mathcal{T} = \{T_1, T_2, ..., T_T\}$ and a set of environments $\mathcal{E} = \{E_1, E_2, ..., E_T\}$ where each task $T_i$ is associated with an environment $E_i$.
Each environment $E_i$ is a Markov Decision Process (MDP) defined by a tuple $(\mathcal{S}, \mathcal{A}, \mathcal{P}, \mathcal{R}, \gamma)$ where $\mathcal{S}$ is the state space, $\mathcal{A}$ is the action space, $\mathcal{P}$ is the transition probability function, $\mathcal{R}$ is the reward function, and $\gamma$ is the discount factor.
The goal of the agent is to learn a policy $\pi_{\theta}$ that maximizes the expected return $J(\pi_{\theta})$ where $\theta$ are the parameters of the policy.

In a multi-task setting, the agent has to learn a policy $\pi_{\theta_i}$ for each task $T_i$. The agent is evaluated on its ability to learn a policy for each task $T_i$ so therefore we are interested in the average return over all tasks. Since we evaluate our method on Meta-World, we will focus on the discrete signal that is the success rate of the agent on each task. Because we consider a multi-task setting with no unseen tasks during training, we do not consider the problem of generalization to unseen tasks for this work. 

\subsubsection{Notation}
\label{sec:notations}
We introduce the following notation:
\begin{itemize}
    \item $I$ the input dimension, 
    \item $O$ the output dimension, 
    \item $H$ the hidden dimension
    \item $D$ the task-specific dimension,
    \item $T$ the number of tasks,
    \item $N$ the number of hidden layers (meaning that we have $N+1$ layers in total).
\end{itemize}
In addition, we note the input $x$ and $x_i$ as the input of the $i$-th layer (e.g., $x = x_0$). Additionally, we define:
\begin{itemize}
    \item $\text{SH}^i$ the shared $i$-th linear layer.
    \item $\text{TS}_j^i$ the $i$-th task specific layer for the $j$-th task.
    \item $P_{\text{down}}^i$ the $i$-th projection layer that projects from $H$ to $D$ (except for $P_{\text{down}}^0$ that projects from $I$ to $D$).
    \item $P_{\text{up}}^i$ the $i$-th projection layer that projects from $D$ to $H$ (except for $P_{\text{up}}^N$ that projects from $D$ to $O$).
\end{itemize}

\subsubsection{Projection to the task-specific dimension}
\label{sec:projection}
In a transformer setup, we often have $I = H = O$, which allows PAL to have a single shared down projection $P_{\text{down}}^*$ and a single shared up projection $P_{\text{up}}^*$. Under a maximum number of parameters constraint, it is better to have a single shared projection than a projection for each layer~\cite{pal}.

This is less relevant in our case as $I$ and $O$ are not necessarily equal to $H$ and the gain of sharing the projection is less important as transformers such as BERT have 12 layers while we only have 3 hidden layers (meaning 4 layers in total). Thus, we experiment with both options: a single shared projection and a projection for each layer. In the case of a single shared projection, we will note $P_{\text{down}}^*$ and $P_{\text{up}}^*$ the shared projection layers. Note that we still have two independent projections $P_{\text{down}}^0$ and $P_{\text{up}}^N$ that are not shared due to the difference in input and output dimensions.

In our case, we investigate whether having these two specific projections is worthwhile. Indeed if $I$ and $O$ are small, it may be more beneficial to skip the projection so that $\text{TS}_j^0$ goes from $I$ to $D$ and $\text{TS}_j^N$ goes from $D$ to $O$.
We can compute the number of parameters for each case:
\begin{itemize}
    \item \textbf{First Task-specific layer $\text{TS}_j^0$}: with a projection the number of parameters is $I \times D + T . (D \times D + D)$ while without a projection it is $T . (I \times D + D)$.
    \item \textbf{Last Task-specific layer $\text{TS}_j^N$}: with a projection the number of parameters is $D \times O + T . (D \times D + D)$ while without a projection it is $T . (D \times O + O)$.
\end{itemize}
For Meta-World, it was more beneficial to have an individual down projection $P_{\text{down}}^0$ but no up projection $P_{\text{up}}^N$ as $I = 104$ (output of the CARE encoder) and $O = 8$.

\subsubsection{Residuals across task-specific layers}
\label{sec:residuals}
We also add a residual connection between the task-specific layers, as illustrated by the dotted red in Figure \ref{fig:ptsl}.
Let us note the function $g: (\mathbb{R}^D, \mathbb{R}^D) \rightarrow \mathbb{R}^D$ that combines the projection of our input $P_{\text{down}}^i(x_i)$ with the output of the previous task-specific layer $\text{TS}_j^i(P_{\text{down}}^{i-1}(x_{i-1}))$ to obtain the input of the current task-specific layer $\text{TS}_j^i$.

In this paper we will consider four different functions $g$:
\begin{itemize}
    \item \textbf{No residual}: $g(x_i^{\text{proj}}, y_{i-1}^{\text{task}}) = x_i^{\text{proj}}$.
    \item \textbf{Addition}: $g(x_i^{\text{proj}}, y_{i-1}^{\text{task}}) = x_i^{\text{proj}} + y_{i-1}^{\text{task}}$.
    \item \textbf{Learnable sum}: $g(x_i^{\text{proj}}, y_{i-1}^{\text{task}}) = \alpha x_i^{\text{proj}} + \beta y_{i-1}^{\text{task}}$ with $\alpha, \beta \in \mathbb{R}$ learnable parameters.
    \item \textbf{Learnable projection}: two vectors are concatenated to a vector of $\mathbb{R}^{2D}$ that is projected to $\mathbb{R}^{D}$.
    $g(x_i^{\text{proj}}, y_{i-1}^{\text{task}}) = P_g \times \text{Concat}\left( x_i^{\text{proj}}, y_{i-1}^{\text{task}} \right)$ with $P_g \in \mathbb{R}^{D \times 2D}$, a learnable projection shared across tasks and layers.
\end{itemize}

\section{Experiments}
\label{sec:experiments}

In this section, we evaluate PTSL in the Meta-World multi-task RL environment and compare against baselines. We also conduct ablation studies to verify the effectiveness of our method. To support the community, our code is made publicly available at \url{https://github.com/JosselinSomervilleRoberts/PTSL} which is a fork of the CARE~\cite{care} repository that can be found at \url{https://github.com/facebookresearch/mtrl}.
In addition to the code, the repository contains all the commands to reproduce our experiments, our run results, and some implementation and training tips. All trainings were performed on a NVIDIA RTX A6000 and averaged across $n$ runs.

The first goal of our experimental evaluation is to assess if the PSTL architecture improves the performance of a multi-task agent without increasing the number of parameters. This comparison is done in two different settings: short horizon to evaluate the sample efficiency of PTSL and long horizon to assess convergence to an efficient policy. We compare our method to current state-of-the-art architecture baselines: CARE~\cite{care}, Soft Modularization~\cite{softmod}, and MT-SAC. The choice of baselines is described in detail in Section \ref{sec:baselines}.


We also assess which variant of PTSL yields the best results, and evaluate whether having a shared or independent projection is better and what residual function should be used. The residual functions considered are no residual, the addition, the learnable sum, and the learnable projection (See Section \ref{sec:residuals} for the definitions). 

\subsection{Benchmark}
\label{sec:benchmark}

\begin{figure}[h]
    \centering
    \includegraphics[width=0.5\textwidth]{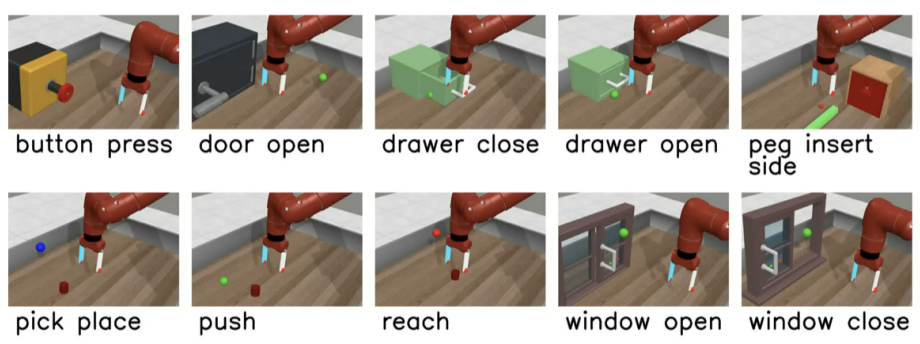}
    \caption{The MT10 benchmark from Meta-World contains 10 tasks: reach, push, pick and place, open door, open drawer, close drawer, press button top-down, insert peg side, open window, and open box.}
    \label{fig:mt10}
\end{figure}

We use Meta-World's MT10 and MT50 Goal-conditioned tasks as our benchmarks. Meta-World is a multi-task RL benchmark containing 50 robotic manipulation tasks performed by a simulated Sawyer robot. MT10 and MT50 are two evaluation protocols based on Meta-World, where MT10 contains 10 tasks (shown in Fig. \ref{fig:mt10}), and MT50 contains all 50 tasks. The state space is 12-dimensional and consists of tuples of 3D Cartesian end-effector position, 3D Cartesian positions of one or two objects, and the goal position. All our tasks in MT10 and MT50 are goal-conditioned tasks. 

\subsection{Baselines}
\label{sec:baselines}

We compare our method to the following baselines that are implemented in the CARE and our repository:

\textbf{MT-SAC}: a shared backbone\cite{sac} baseline with disentangled alphas~\cite{garage} that has a simple shared fully connected backbone. There are $N=3$ hidden layers with a hidden dimension of $H=400$ resulting in $P=1,641,222$ parameters for both MT10 and MT50.

\textbf{Soft Modularization}: a soft modularization baseline that learns different policies for each task using a routing network. With also $N=3$ hidden layers with a hidden dimension of $H=400$ resulting in $P=485,766$ parameters for MT10 and $P=485,806$ for MT50.

\textbf{CARE}: the CARE~\cite{care} baseline\footnotemark{}. Also $N=3$ hidden layers with a hidden dimension of $H=400$. There are $A=4$ experts in the Mixture of Experts with a hidden dimension of $H_A = 50$ resulting in $P=1,871,534$ parameters for both MT10 and MT50.
\footnotetext{Although we used the implementation provided by CARE with the same parameters and the same version of Meta-World, we were unable to reproduce their results even after averaging over 10 or 20 seeds. This is a known issue (see\url{https://github.com/facebookresearch/mtrl/issues}). The results provided here may appear different from what was published in the original paper. In the rest of this work, for the purpose of comparison, all the reported CARE values will be the results that we were able to reproduce on the same horizon and not the results published in the original paper.}

\textbf{CARE + PTSL \textit{(ours)}}: The proposed method on top of CARE. To keep the comparison fair, we use nearly the same number of parameters and layers as CARE, with $P=1,871,532$ parameters and $N=3$ layers. To obtain these numbers, we had to reduce $H$ to $326$ and set $D$ to $50$ for MT10 (with shared projections) and we kept the same encoder parameters as CARE. For MT50, we reduced $D$ to $32$ and $H$ to $274$  to reach $P=1,869,908$ parameters.
    
\textbf{CARE + PTSL (shallow) \textit{(ours)}}: Variation of PTSL with fewer layers on top of CARE, $N=2$ hidden layers, $H=400$, and $D=50$ resulting in $P=1,556,454 $ parameters. All other parameters are identical to the previous CARE + PTSL architecture for MT10. We show that with fewer layers and less parameters, we obtain similar performance on a short horizon. This method is only tested for MT10.
    
\textbf{PTSL only \textit{(ours)}}: The proposed PTSL as a simple standalone shared backbone with disentangled alphas. Once again we set $N=3$ and we chose $H$ and $D$ to match the number of parameters of CARE. This means that for MT10, $H=367$ and $D=50$ resulting in $P=1,869,833$ parameters. For MT50, $H=325$ and $D=32$ resulting in $P=1,871,187$ parameters.

\subsection{Comparative evaluation}
\label{sec:results}

\begin{figure*}[]
  \captionsetup[subfigure]{justification=centering, labelformat=empty} 
  \centering
  
  \begin{subfigure}{0.245\linewidth}
    \includegraphics[width=\linewidth]{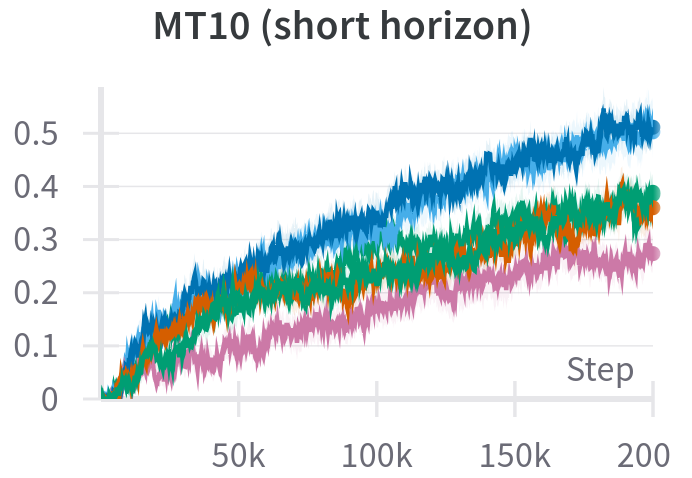}
    \caption{(a)}
    \label{fig:mt10-results_short_horizon}
  \end{subfigure}
  \begin{subfigure}{0.245\linewidth}
    \includegraphics[width=\linewidth]{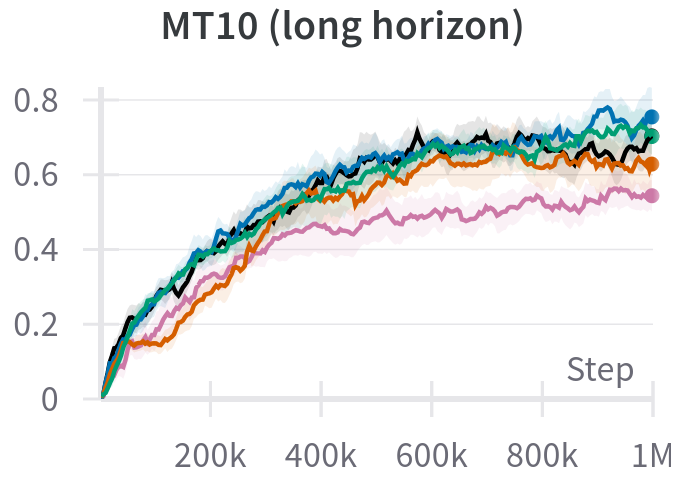}
    \caption{(b)}
    \label{fig:mt10-results_long_horizon}
  \end{subfigure}
  \begin{subfigure}{0.245\linewidth}
    \includegraphics[width=\linewidth]{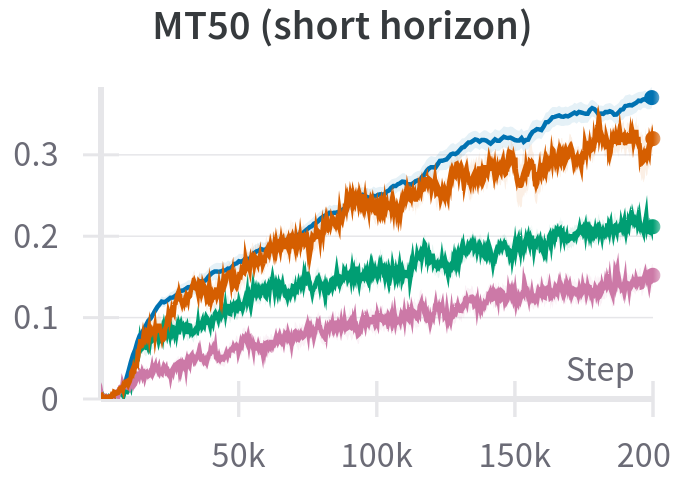}
    \caption{(c)}
    \label{fig:mt50-results_short_horizon}
  \end{subfigure}
  \begin{subfigure}{0.245\linewidth}
    \includegraphics[width=\linewidth]{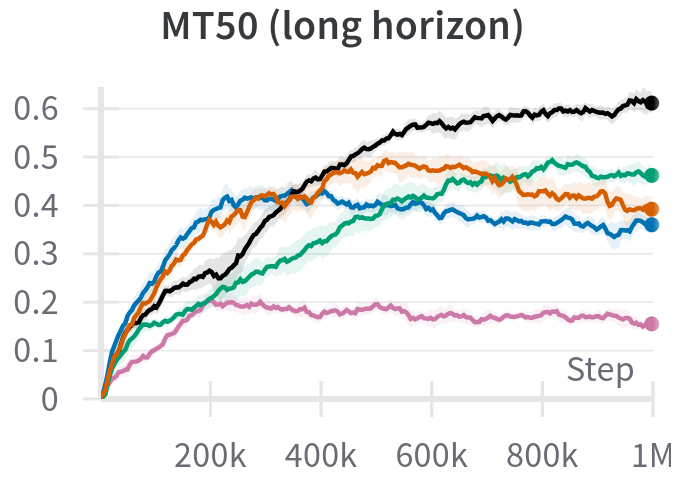}
    \caption{(d)}
    \label{fig:mt50-results_long_horizon}
  \end{subfigure}
  \begin{subfigure}{1.0\linewidth}
    \includegraphics[width=\linewidth]{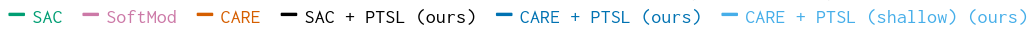}
  \end{subfigure}

  \caption{Training curves of different methods on all benchmarks. For MT10, PTSL converges faster than baselines, and for MT50, we see a gain in sample efficiency. The bolded line represents the mean over $n=10$ runs for the short horizon and $n=4$ for the long horizon. The shaded area represents the standard error. }
  \label{fig:curves}
\end{figure*}

\begin{table}[!h]
    \centering
    \scalebox{1.0}{
    \begin{tabular}{ l@{\hspace{0.4cm}}c@{\hspace{0.5cm}}c@{\hspace{0.5cm}}c}
      \toprule
      Success | 200K ($n=10$) & MT10-Cond. & MT50-Cond. \\
      \midrule
      Multitask SAC~\cite{metaworld}& 0.389 ± 0.029 & 0.212 ± 0.018\\
      Soft Mod.~\cite{softmod} & 0.274 ± 0.033 & 0.152 ± 0.022\\
      CARE~\cite{care}& 0.360 ± 0.030 &  0.320 ± 0.027 \\
      \textbf{CARE + PTSL (Ours)} & \textbf{0.511} ± \textbf{0.034} & \textbf{0.370} ± \textbf{0.020}\\
      CARE + PTSL (Shallow) (Ours) & 0.503 ± 0.051 & - \\
      \bottomrule
    \end{tabular}
}
    \caption{Success rate of the baselines on the short horizon (200k steps per task) for MT10 and MT50 Goal-conditioned benchmark. Results are averaged over $n=10$ seeds for each method. We report the mean and standard error.}
    \label{tab:comparison_baselines}
\end{table}

\begin{table}[!h]
    \centering
    \scalebox{1.0}{
    \begin{tabular}{ l@{\hspace{0.4cm}}c@{\hspace{0.5cm}}c@{\hspace{0.5cm}}c}
      \toprule
      Success | MT10-Cond. & After 1M steps & Best \\
      \midrule
      Multitask SAC~\cite{metaworld}& 0.706 ± 0.050 & 0.737 ± 0.055\\
      Soft Mod.~\cite{softmod} & 0.533 ± 0.039 & 0.554 ± 0.050\\
      CARE~\cite{care}& 0.648 ± 0.060 &  0.683 ± 0.066\\
      \textbf{CARE + PTSL (Ours)} & \textbf{0.742} ± \textbf{0.067} & \textbf{0.772} ± \textbf{0.053}\\
      PTSL only (Ours) & 0.697 ± 0.043 & 0.721 ± 0.050\\
      \bottomrule
    \end{tabular}
    }
    \caption{Success rate of the baselines on MT10 Goal-conditioned on the long horizon (1M steps per task). Results are reported both at the end of the 1M steps and at the best average value. The Results are averaged over $n=4$ seeds for each method. We report the mean and standard error.}
    \label{tab:mt10-long_comparison_baselines}
\end{table}

\begin{table}[!h]
    \centering
    \scalebox{1.0}{
    \begin{tabular}{ l@{\hspace{0.4cm}}c@{\hspace{0.5cm}}c@{\hspace{0.5cm}}c}
      \toprule
      Success | MT50-Cond. & After 1M steps & Best \\
      \midrule
      Multitask SAC~\cite{metaworld}& 0.466 ± 0.013 & 0.489 ± 0.016\\
      Soft Mod.~\cite{softmod} & 0.154 ± 0.010 & 0.206 ± 0.018\\
      CARE~\cite{care}& 0.388 ± 0.028 &  0.495 ± 0.024\\
      CARE + PTSL (Ours) & 0.354 ± 0.015 & 0.427 ± 0.020\\
      \textbf{PTSL only (Ours)} & \textbf{0.610} ± \textbf{0.021} & \textbf{0.614} ± \textbf{0.020}\\
      \bottomrule
    \end{tabular}
    }
    \caption{Success rate of the baselines on MT50 Goal-conditioned on the long horizon (1M steps per task). Results are reported both at the end of the 1M steps and at the best average value. Results are averaged over $n=4$ seeds for each method. We report the mean and standard error.}
    \label{tab:mt50-long_comparison_baselines}
\end{table}

Figure \ref{fig:mt10-results_short_horizon} shows the average success rate on the 10 tasks of the MT10 Goal-conditioned benchmark from Meta-World\cite{metaworld} for MT-SAC, Soft Modularization, CARE, and CARE~+~PTSL (both deep and shallow architectures). Success rate is noisy since it is a binary variable, so we averaged the results across multiple seeds (the number of seeds is noted as $n$, set to 10 for short horizon and 4 for long horizon).

We consider 1 million steps as our long horizon, and 200 thousand steps as our short horizon. As noted earlier, all methods are trained using SAC with disentangled alphas.

As explained in Section \ref{sec:baselines}, CARE does not perform as well as described in the original paper~\cite{care}. In particular, we noticed decreasing performances on long horizons (See Figure \ref{fig:mt50-results_long_horizon}). To remove uncertainty coming from CARE in our experiments, we trained two PTSL agents on long horizons, one with CARE and one without.

Table \ref{tab:comparison_baselines} and Fig. \ref{fig:mt10-results_short_horizon} show that our method outperforms all baselines on the short horizon on MT10. For comparison, in the Meta-World paper~\cite{metaworld}, it takes around 1.5M steps for the Multitask SAC agent to reach the accuracy that our CARE + PTSL agent reaches within 200K steps. This suggests that our method is highly sample-efficient. Furthermore, we show that using a shallow PTSL network yields very similar results on MT10 after a short horizon, suggesting that lighter PTSL architectures can still generalize. 

Table \ref{tab:comparison_baselines} and Figure \ref{fig:mt50-results_short_horizon} also show that PTSL is beneficial as CARE + PTSL outperforms all baselines with more tasks (MT50) even if this means reducing the size of the hidden layers (to keep the same number of parameters).

Tables \ref{tab:mt10-long_comparison_baselines} and \ref{tab:mt50-long_comparison_baselines} as well as Figures \ref{fig:mt10-results_long_horizon} and \ref{fig:mt50-results_long_horizon} show that PTSL is able to learn a good policy on the long horizon. For MT10, CARE + PTSL performs best and reaches a top success rate of $0.772$. For MT50, MT-SAC was the best-performing agent since CARE stopped learning after 400K steps. That is why while the CARE + PTSL agent does not perform exceptionally, the standalone PTSL outperforms all methods. Importantly, PTSL achieves a \textbf{score of $\mathbf{0.61}$ on MT50 Goal-Conditioned after only 1 million steps}. This surpasses the reported results from the original CARE paper\cite{care} of $0.54$ after 2 million steps, and the Soft-Modularization paper\cite{softmod} of $0.60$ after 1 million steps \textbf{on MT50-Fixed} (which is easier than Goal-Conditioned).

In all settings, adding PTSL to the best-performing method improved the performance, whether it was CARE or a simple Multitask SAC, suggesting PTSL can improve any method. 

One issue while training a PTSL architecture is to balance the training of the shared layers and the task-specific ones. Ideally, a shared policy should be learned first and then the task-specific layers should be learned to improve the performance. Our experiments showed that initializing the projection matrices to zero was a good way to enforce this.

\subsection{Ablation study}
\label{sec:ablation}

\begin{table}[!h]
    \centering
    \scalebox{1.0}{
    \begin{tabular}{ l@{\hspace{0.4cm}}c@{\hspace{0.5cm}}c@{\hspace{0.5cm}}c}
      \toprule
      Success | MT10-Cond. & 200K ($n=10$)\\
      \midrule
      Independent projection & 0.369 ± 0.045\\
      \textbf{Shared projection} & \textbf{0.511} ± \textbf{0.034} \\
      \bottomrule
    \end{tabular}
    }
    \caption{Success rate of CARE + PTSL with independent versus shared projections on MT10 Goal-Conditioned. Results are averaged over $n=4$ seeds for each method. We report the mean and standard error.}
    \label{tab:comparison_projection}
\end{table}

\begin{table}[!h]
    \centering
    \scalebox{1.0}{
    \begin{tabular}{ l@{\hspace{0.4cm}}c@{\hspace{0.5cm}}c@{\hspace{0.5cm}}c}
      \toprule
      Success | MT10-Cond. & 200K ($n=10$)\\
      \midrule
      \textbf{No residual} & \textbf{0.511} ± \textbf{0.034}\\
      Learnable sum & 0.410 ± 0.031 \\
      Learnable projection & 0.385 ± 0.032 \\
      \bottomrule
    \end{tabular}
    }
    \caption{Success rate of CARE + PTSL with various residual functions. Results are averaged over $n=4$ seeds for each method. We report the mean and standard error. The Sum residual is not indicated as it did not converge.}
    \label{tab:comparison_residual}
\end{table}

\begin{figure}[h]
  \captionsetup[subfigure]{justification=centering, labelformat=empty} 
  \centering
  
  \begin{subfigure}{0.49\linewidth}
    \includegraphics[width=\linewidth]{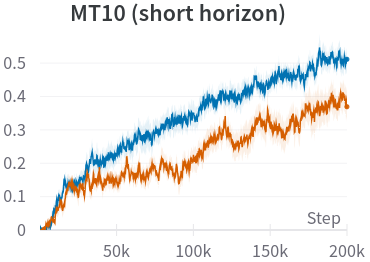}
    \caption{(a)}
    \label{fig:shared_vs_independent}
  \end{subfigure}
  \begin{subfigure}{0.49\linewidth}
    \includegraphics[width=\linewidth]{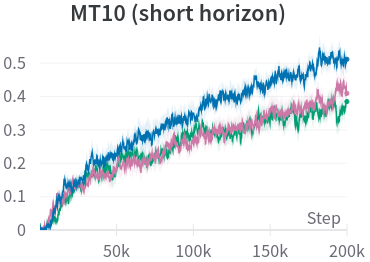}
    \caption{(b)}
    \label{fig:residual}
  \end{subfigure}
  \begin{subfigure}{1.0\linewidth}
    \centering
    \includegraphics[width=0.85\linewidth]{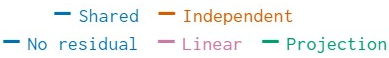}
  \end{subfigure}

  \caption{Training curves of different methods on all benchmarks. The bolded line represents the mean over $n=10$ runs for the short horizon and $n=4$ for the long horizon. The shaded area represents the standard error.}
  \label{fig:curves2}
\end{figure}

The result in the previous section suggests that PTSL (with CARE or standalone) is both sample efficient and yields a good policy on long horizons. In this section, we further examine the influence of the different components of the PTSL architecture on the performance of the model. 
\\

\textbf{Shared versus independent projections.} In Section \ref{sec:projection}, we discussed the relevance of using a shared projection in our context since we have only a small number of layers.
We decided to compare the two approaches on the short horizon setting with the same parameters: $H=326$, $D=50$, and $N=3$. This means that the independent projection has slightly more parameters (about 12\% more).
We show the results in Figure \ref{fig:shared_vs_independent}.

Figure \ref{fig:shared_vs_independent} shows that the shared projection is better than the independent projection, and this is the case even though the independent projection has more parameters.
The shared projection is therefore more efficient than the independent projection in our context, as it helps the network to learn a consistent mapping between the shared embedding space and the task-specific embedding space.
This result is in agreement with the results from Stickland et. al.~\cite{pal} that show the shared projection is better than the independent projection for NLP. The shared projection is also more stable than the independent projection as it has a lower variance.

\textbf{Comparison of residual functions.} In Section \ref{sec:residuals}, we discussed the relevance of using residual functions as they have shown great results in other Deep Learning tasks\cite{residual}. To verify if residuals are relevant for PTSL, we evaluated four variants on the short horizon setting with the same number of parameters.
We used $H=326$, $D=50$, and $N=3$ for all methods except the \textit{Learnable projection} that uses $H=321$ resulting in roughly $1.871$ million parameters.

Figure \ref{fig:residual} and Table \ref{tab:comparison_residual} shows that having no residuals is better on the short horizon. This means that while the model is more expressive, the additional complexity makes it less sample-efficient and is not beneficial.

\section{Conclusion}
\label{sec:conclusion}

In this work, we present the \textbf{Projected Task-Specific Layers (PTSL)}, a novel method inspired by NLP, that surpasses the state-of-the-art on the MT10 and MT50 Goal-Conditioned benchmark from Meta-World.

Here we showed that PTSL learns a high-performing policy faster than other popular methods with higher sample efficiency and without introducing more parameters. Furthermore, PTSL can be integrated \textit{with} existing methods like CARE to improve them. Finally, we showed the benefits of sharing a low-dimensional embedding space between the shared backbone and the task-specific layers, which becomes more obvious with task diversity (MT50). 

In future work, we will include transfer learning of individual layers from MT10 to MT50, hierarchical individual layers to better scale, and the implementation of a routing network for the individual layers to better share parameters.


\section*{ACKNOWLEDGMENT}
The authors would like to thank Saurabh Kumar, Yoni Gozlan, Paul-Emile Giacomelli and Brian Park for their feedback and advice during the preparation of this manuscript.


\bibliographystyle{IEEEtran}
\bibliography{IEEEabrv,egbib}

\end{document}